\documentclass{article}

\PassOptionsToPackage{numbers, compress}{natbib}
    \usepackage[preprint]{neurips_2024}

\usepackage[utf8]{inputenc} 
\usepackage[T1]{fontenc}    
\usepackage{hyperref}       
\usepackage{url}            
\usepackage{booktabs}       
\usepackage{amsfonts}       
\usepackage{nicefrac}       
\usepackage{microtype}      
\usepackage{xcolor}         

\usepackage{graphicx} 
\usepackage{algorithm}
\usepackage{algorithmic}
\usepackage{graphicx}
\usepackage{amsmath}
\usepackage{amssymb}
\usepackage{cleveref}
\usepackage{subcaption}
\usepackage{wrapfig}

\usepackage{bm}
\title{Probability Passing for Graph Neural Networks: Graph Structure and Representations Joint Learning}

\author{
  Ziyan Wang$^{1}$, Yaxuan He$^{2}$, Bin Liu$^{1}$\\
  $^1$ School of Statistics, Southwestern University of Finance and Economics, Chengdu, China\\
  $^2$ Department of Statistics, Rutgers University, Piscataway, NJ 08854\\
  \texttt{wangzy187@163.com, yaxuan.he@rutgers.edu, liubin@swufe.edu.cn}
}

\begin{document}

\maketitle

\begin{abstract}
Graph Neural Networks (GNNs) have achieved notable success in the analysis of non-Euclidean data across a wide range of domains. However, their applicability is constrained by the dependence on the observed graph structure. To solve this problem, Latent Graph Inference (LGI) is proposed to infer a task-specific latent structure by computing similarity or edge probability of node features and then apply a GNN to produce predictions. 
Even so, existing approaches neglect the noise from node features, which affects generated graph structure and  performance. In this work, we introduce a novel method called Probability Passing to refine the generated graph structure by aggregating edge probabilities of neighboring nodes based on observed graph. Furthermore, we continue to utilize the LGI framework, inputting the refined graph structure and node features into GNNs to obtain predictions. We name the proposed scheme as Probability Passing-based Graph Neural Network (PPGNN). Moreover, the anchor-based technique is employed to reduce complexity and improve efficiency. Experimental results demonstrate the effectiveness of the proposed method. Code can be found at \url{https://github.com/JiyaWang/PPGNN}.

\end{abstract}

\section{Introduction}

Graph neural networks (GNNs) have gained considerable attention in recent years for their efficiency in analyzing graph data across various real-world applications \citep{defferrard2016convolutional,kipf2017semi,hamilton2017inductive,veli2018graph}, such as microscopic molecular networks \cite{li2018adaptive}, protein networks \cite{strokach2020fast}, as well as macroscopic social networks, traffic networks \cite{wang2020traffic}, and industrial chain \cite{ijcai2023p674}. In the aforementioned scenarios, it is assumed that the observed graph structure is optimal for downstream task. However, the observed graph may be full of noise~\cite{franceschi2019learning} and incompleteness, stemming from errors in data collection or measurement. 
For instance, in studies on the toxicity of compounds, it is challenging to regress the toxicity of compounds based on molecular graphs~\cite{li2018adaptive}. 
Similarly, in the domain of macroscopic social networks, connections among individuals on social media may not effectively reflect their underlying relevance.

To this end, Latent Graph Inference (LGI) ~\cite{kazi2022differentiable, chen2020iterative, fatemi2021slaps, wu2022nodeformer} is proposed to jointly learn the latent graph and corresponding node embeddings using graph learner and GNNs. Specifically, the graph learner first computes the similarity or edge probabilities between node pairs , and then generates graph structure via sampling from edge distribution. Subsequently, GNNs take the node features and the generated graph as input to produce the predictions for downstream tasks. LGI plays a vital role in improving predictive performance by rectifying the process of aggregating neighbor information~\cite{topping2022understanding,kazi2022differentiable}. Unfortunately, there exist two weaknesses that may limit its performance. Firstly, these models assume that the node features are noiseless and that graph noise only arises from graph topology, neglecting the impact of noise from node features. Secondly, it's expensive to compute similarity or probabilities for all node pairs.

\begin{figure}[t]
\centering
\begin{minipage}[t]{0.3\textwidth}
\centering
\includegraphics[width=1\textwidth]{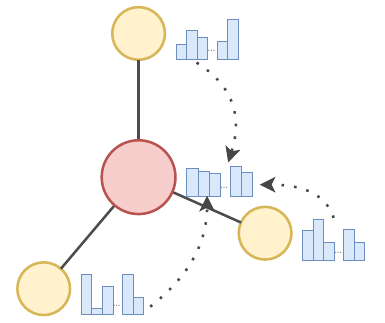}
\caption{A sketch of the proposed Probability Passing. The blue bar charts represent the edge probability distributions. The red node and yellow nodes represent the central node and its neighbors respectively. }
\label{fig:ProbabilityPassing}
\end{minipage}
\hfill
\begin{minipage}[t]{0.65\textwidth}
\centering
\includegraphics[width=1\textwidth]{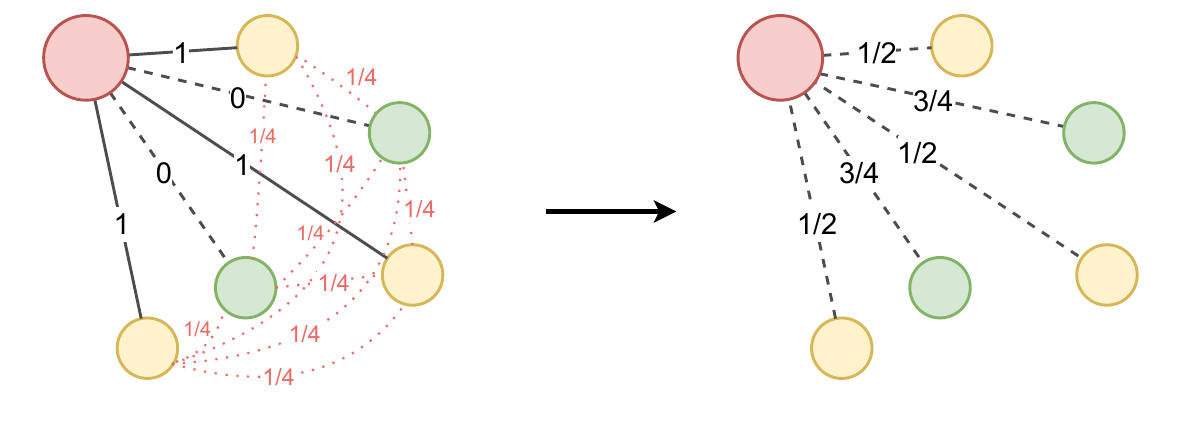}
\caption{The interpretation of Probability Passing from the perspective of a two-step transition probability. The observed adjacency matrix can be viewed as a 0-1 probability matrix. In this example, the generated edge probability distribution is uniform, and the green nodes represent nodes that are not connected to the central node. The final probability from the central node to the target node is obtained through a special form of two-step transition probabilities generated by the adjacency matrix and the generated probability matrix.}
\label{fig:explainProbabilityPassing}
\end{minipage}
\end{figure}

To address the first issue, we propose an innovative approach named Probability Passing, similar to Message Passing. As illustrated in \cref{fig:ProbabilityPassing}, this method updates the probability distribution of the central node with other nodes by aggregating the probability distributions of neighboring nodes along the observed graph structure, which means that if two nodes are directly connected in the observed graph, the edge probabilities 
between the pairs of them and a third node will be similar. It allows to correct the inaccurate edges and add the missing edges in the generated graph structure. In addition, we can explain Probability Passing from the respective of transition probability as shown in \cref{fig:explainProbabilityPassing}. Typically, when computing two-step transition probabilities, two identical probability matrices are used. In the case of Probability Passing, the adjacency matrix can be treated as a 0-1 probability matrix. Probability transfer is performed on the adjacency matrix, and then on the generated probability matrix, resulting in the regeneration of the probability matrix.

To address the second problem, we develop anchor-based method~\cite{chen2020iterative, Liu_He_Chang_2010, Wu_Yen_Zhang_Xu_Zhao_Peng_Xia_Aggarwal_2019}. Specifically, we initially randomly sample from nodes to produce anchors. Subsequently, for every node, we calculate the edge probability between this node and the corresponding anchors. Leveraging this probability, we differentiably sample the edges to obtain the generated graph structure. 
The anchor-based method not only reduces the time and memory complexity, but also improves efficiency. Intuitively, anchors play as message relay stations, responsible for receiving information and passing it.


To summarize, we outline the main contributions of this paper as follows:
\begin{itemize}
    \item We propose a novel Probability Passing method, which updates the edge probability distribution of central nodes by aggregating edge probabilities of neighbors along the observed graph structure.
    \item The anchor-based method presented in this paper offers an efficient approach reducing both time and memory complexity while improving efficiency by utilizing anchors as message relay stations.
    \item We have validated the effectiveness and robustness of the model on real-world datasets.
\end{itemize}

\section{Related Work}

GNNs have emerged as crucial tools for processing structured graph data, and achieved significant success in various domains such as social networks~\citep{guo2020deep}, bioinformatics~\cite{li2021graph} and recommendation systems~\citep{wu2022graph}. Traditional GNNs are typically used under the assumption of a complete and accurate graph structure. However, in practical applications, one of the challenges is that the initial graph may be unavailable, incomplete, or full of noise.


In scenarios where the initial graph is not given or incomplete, people have to dynamically extract structural information from the features. This requires modeling the connectivity patterns of the graph during the inference process. For instance, \cite{wang2019dynamic} proposed the EdgeConv model to handle point cloud data, which lacks explicit edge information. EdgeConv combined point cloud networks and edge convolutions to construct the graph structure. 
Similarly, ~\cite{franceschi2019learning} proposed to learn latent discrete structures by jointly learning the graph structure and GNN parameters through two levels of optimization problems. The outer-level problem aims to learn the graph structure, and the inner-level problem involves optimizing model parameters in each iteration of the outer-level problem. This approach successfully handles the application of GNNs in scenarios with incomplete or damaged graphs.


In scenarios where the observed graph data contains noise, graph fusion including multi-modal graph fusion \cite{10058089,ektefaie2023multimodal,wei2023mm,cai2022multimodal,mai2020modality} and fusing observed graphs with inference graphs \cite{chen2020iterative,sun2023self,kazi2022differentiable,de2022Latent,pan2023beyond} prove to be effective methods.

Multi-modal methods have demonstrated the potential to learn an accurate graph based on multiple observed graphs.
For example, \cite{10058089} proposed the Multimodal Fusion Graph Convolutional Network (MFGCN) model to extract spatial patterns from geographical, semantic, and functional relevance, which has been applied in accurate predictions for online taxi services.
~\cite{ektefaie2023multimodal} presented multi-modal graph AI methods that combine different inductive preferences and leverage graph processing for cross-modal dependencies. 
~\cite{wei2023mm} proposed a graph neural network model to fuse two-modal brain graphs based on diffusion tensor imaging (DTI) and functional magnetic resonance imaging (fMRI) data for the diagnosis of ASD (Autism Spectrum Disorder). 
In \cite{cai2022multimodal}, the authors consider continual graph learning by proposing the Multi-modal Structure-Evolving Continual Graph Learning (MSCGL) model, aiming to continually adapt their method to new tasks without forgetting the old ones.
\cite{pan2023beyond} utilized graph reconstruction in both feature space and structural space for clustering, effectively addressing the challenges associated with handling heterogeneous graphs.
The aforementioned multi-modal methods ignore potential modal discrepancies. To address this issue, \cite{mai2020modality} proposed an adversarial encoder-decoder-classifier to explore interactions among different modalities.

Another feasible approach is to learn a latent graph and integrate it with the initial graph. For instance, \cite{chen2020iterative} introduced the Iterative Deep Graph Learning (IDGL) model to learn an optimized graph structure. 
\cite{sun2023self} proposed a Graph Structure Learning framework guided by the Principle of Relevant Information (PRI-GSL) to identify and reveal hidden structures in a graph. \cite{kazi2022differentiable} introduced the Differentiable Graph Module (DGM), a learnable function that predicted edge probabilities in a graph, enabling the model to dynamically adjust the graph structure during the learning process. Furthermore, \cite{de2022Latent} generalized the Discrete Deep Generative Model(dDGM) for latent graph learning. 

\section{Method}
\subsection{Preliminary and Problem Definition}
We denote a graph as $\mathcal{G}=(\mathcal{V},\mathcal{E})$ where the node set $\mathcal{V}$ comprises $N$ nodes, and the edge set $\mathcal{E}=$ $\{(i,j)\mid a_{ij}=1\}$ is defined by an initial noisy adjacency matrix $\mathbf{A}^{(0)}=[a_{ij}]_{N\times N}$, where $a_{ij}= 1$ if nodes $i$ and $j$ are connected, and 0 otherwise. Each node has an input feature $\mathbf{x}\in\mathbb{R}^D$ and a label $\mathbf{y}.$ This paper primarily centers on the exploration of graph representation learning in the context of noisy graph structure and features. In this setting,
given a noisy graph input $\mathcal{G} := \{\mathbf{A}^{(0)}, \mathbf{X}\}$, the graph structure learning problem  is to produce an optimized graph $\mathcal{G}^{*}:=\{\mathbf{A}^{(*)}, \mathbf{X}\}$ and node embeddings $\mathbf{H} = \text{GNN}(\mathcal{G}^*,\mathbf{X})$, with respect to the  downstream tasks.
\begin{figure}[t]
\centering
   \includegraphics[width=0.8\linewidth]{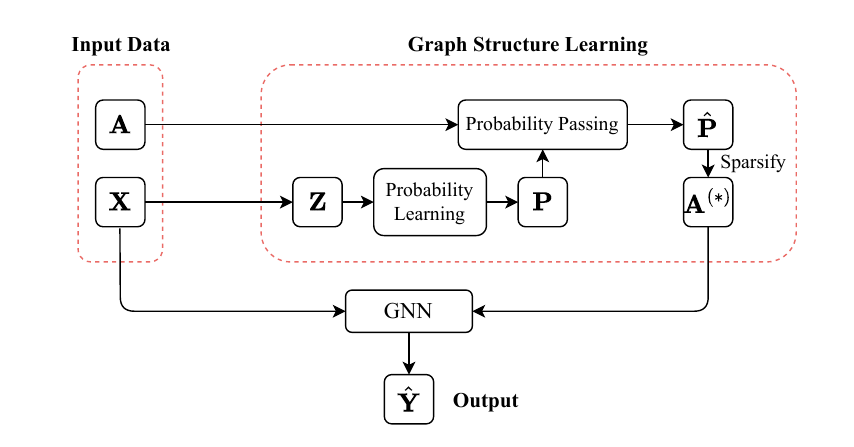}
   \caption{Illustration of the proposed modcel and its data flow. The input graph data entails node features $\mathbf{X}$ and graph adjacency $\mathbf{A}^{(0)}$. The entire model is divided into two parts: graph structure learning and node representations learning. In the graph structure learning process \cref{Graph Learning}, Probability Learning receives $\mathbf{X}$ and $\mathbf{A}^{(0)}$ to produce the probability matrix $\mathbf{P}$. Through probability passing, the probability matrix and the original adjacency matrix are fused, and finally $\mathbf{A}^{(*)}$ is obtained after sparsification. The node representations learning process in \cref{Representation Learning} uses GCN as message-passing method to generate predictions.}
   \label{model}
\end{figure}
\subsection{Network Architecture}
PPGNN consists of two components: 1) graph learning, and 2) node representation learning. \autoref{model} illustrates these components. In the next two subsections, we will explain them. Lastly we describe how graph structure and node representations can be jointly learned.

\subsection{Probability Passing for Graph Learning}
\label{Graph Learning}
Graph structure learning aims to jointly learn the graph structure and corresponding embeddings. Given node features, graph structure learning first models the edge probabilities between node pairs. 
Previous methods construct edge probabilities solely using node features. However, if there is noise in the node features, it will affect the probability matrix. To address this issue, we use the observed graph to correct the probability metric, increasing the probability of edges that truly exist and decreasing the probability of edges that do not.

\textbf{Node-node probability learning.} Common options include gaussian kernel ~\cite{kazi2022differentiable} and attention mechanism \cite{Jiang_Zhang_Lin_Tang_Luo_2019,Vaswani_Shazeer_Parmar_Uszkoreit_Jones_Gomez_Kaiser_Polosukhin_2017}. Similar to \cite{kazi2022differentiable}, we use gaussian kernel as stated in \cref{eq:probability}.
\begin{equation}
    \mathbf{z}_i = \text{f}_{\Phi}(\mathbf{x}_i) \text{,} \quad
    \mathbf{p}_{ij} = e^{-\|\mathbf{z}_i- \mathbf{z}_j\|^2/t}\text{,}
    \label{eq:probability}
\end{equation}
where the function $\text{f}_{\Phi}$, a neural layer mapping input into the latent space, can be a GNN or shallow MLP. Here,
$\mathbf{\Phi}$ and $t$ are learnable parameters, $\| \cdot \|$ is $L$-2 norm. Furthermore, to mitigate the impact of noise in the features on the results, we adopt a novel approach called \textbf{probability passing} to refine the probability metric. Similar to message passing, it aggregates the probabilities of neighboring nodes to the central node on $\mathbf{A}^{(0)}$. We define probability passing as:
\begin{equation}
    \hat{\mathbf{P}} = \mathbf{D}^{-1}\mathbf{A}^{(0)}\mathbf{P}\text{,}
    \label{eq:node probability}
\end{equation}
where $\mathbf{P} \in \mathbb{R}^{N \times N}$ is a matrix composed of elements $\mathbf{p}_{ij}$ and $\mathbf{D}$ is the degree matrix of $\mathbf{A}^{(0)}$.

\textbf{Node-anchor probability learning.}
The above probability metric learning function like \cref{eq:probability} computes edge probabilities for all pairs of graph nodes, which requires $\mathcal{O}(N^2)$ complexity for both computational time and memory consumption. To address this issue, we design the \textbf{anchor-based probability metric learning} technique, which learns a node-anchor probability matrix $\mathbf{R} \in \mathbb{R}^{N \times s}$ ($s$ is the number of anchors and hyperparameter), inspired by \cite{chen2020iterative}. This process only consumes $\mathcal{O}(Ns)$, achieving linear complexity.

Specifically, we first randomly sample $s$ anchors from $\mathcal{V}$ to form the set $\mathcal{U}$. Therefore, the Node-anchor probability matrix $\mathbf{R}$ can be computed by 
\begin{equation}
    \mathbf{r}_{ij} = e^{-\|\mathbf{z}_i- \mathbf{z}_u\|^2/t}\text{,}
    \label{eq:anchor-embed}
\end{equation}
where $\mathbf{z}_u$ is the embedding of anchor node. To further correct probability matrix, we use probability passing:
\begin{equation}
    \hat{\mathbf{R}} = \mathbf{D}^{-1}\mathbf{A}^{(0)}\mathbf{R}\text{.}
    \label{eq:anchor probability}
\end{equation}

\textbf{Graph sparsification via Gumble-Top-$k$ trick.}
In the real world, the adjacency matrix of a graph is generally sparse. However, $\hat{\mathbf{P}}$ from \cref{eq:node probability} and $\hat{\mathbf{R}}$ from \cref{eq:anchor probability} are dense, which not only lowers computational efficiency but also degrades performance. Therefore, we employ the Gumbel-Top-$k$ trick \cite{kool2019stochastic} to sparsify $\hat{\mathbf{P}}$ or $\hat{\mathbf{R}}$ to obtain a new adjacency matrix $\mathbf{A}^{(*)}$.

Specifically, for each node $i$, we extract $k$ edges as the first $k$ elements of $\log(\mathbf{s}_{i})-\log(-\log(\mathbf{q}))$, where $\mathbf{q}\in\mathbb{R}^N$ is uniformly independently distributed in the interval $[0, 1]$ and $\mathbf{s}_i$ represents the probability scores vector for node $i$ and other nodes(or anchors), i.e.,the $i$-th row of $\hat{\mathbf{P}}$ or $\hat{\mathbf{R}}$.
\label{chapter:graph learning}
\subsection{Node Representation Learning}
\begin{algorithm}[!t]
   \caption{General Framework For PPGNN}
   \label{alg:model}

\begin{algorithmic}
   \STATE {\bfseries Input:}  Node feature matrix $\mathbf{X}$, adjacency matrix $\mathbf{A}^{(0)}$
   \STATE {\bfseries Output:} Predicted node labels $\hat{\mathbf{y}}$
   \IF{PPGNN}
   \STATE $\mathbf{P} \leftarrow \text{ProbabilityLearning}(\mathbf{X},\mathbf{A}^{(0)}$) using 
 \cref{eq:probability}
 \STATE $\hat{\mathbf{P}} \leftarrow \text{ProbabilityPassing}(\mathbf{A}^{(0)},\mathbf{P}$) using \cref{eq:node probability}
   \STATE $\mathbf{A}^{(*)} \leftarrow \text{Gumble-top-}k(\hat{\mathbf{P}})$

   \FOR{$l \in \{0,1,\cdots,L-1\}$}
        \STATE $\mathbf{U}^{(l)} \leftarrow \text{GNN}( \mathbf{A}^{(*)}, \mathbf{U}^{(l-1)};\mathbf{W}^{(l)})$
    \ENDFOR 
    \ELSIF{PPGCN-anchor}
    
       \STATE $\mathbf{R} \leftarrow \text{ProbabilityLearning}(\mathbf{X},\mathbf{A}^{(0)}$) using 
 \cref{eq:anchor-embed}
 \STATE $\hat{\mathbf{R}} \leftarrow \text{ProbabilityPassing}(\mathbf{A}^{(0)},\mathbf{R}$) using \cref{eq:anchor probability}
   \STATE $\mathbf{A}^{(*)} \leftarrow \text{Gumble-top-}k(\hat{\mathbf{R}})$ 

   \FOR{$l \in \{0,1,\cdots,L-1\}$}
        \STATE $\mathbf{V}^{(l)} \leftarrow \text{MP}_1(\mathbf{U}^{(l-1)},\mathbf{A}^{(*)})$ using \cref{eq:the first step}
        \STATE $\mathbf{U}^{(l)} \leftarrow \text{MP}_2( \mathbf{V}^{(l)}, \mathbf{A}^{(*)};\mathbf{W}^{(l)})$ using \cref{eq: the second step}
    \ENDFOR 
   
    \ENDIF
    \STATE $\hat{\mathbf{y}} \leftarrow \text{MLP}(\mathbf{U}^{(L)})$
    \STATE Update $\{\mathbf{\Phi},t,\mathbf{W}^{(l)}\}$ with $\mathcal{L}$ using \cref{chapter:loss}
    
\end{algorithmic}
\end{algorithm}
\label{Representation Learning}
Node representation learning and graph learning are two separate processes that do not interfere with each other. In this paper, we utilize an $L$-layer GCN to learn node features, which accepts $\mathbf{A}^{(*)}$ in \cref{chapter:graph learning} and $\mathbf{X}$ as inputs,
\begin{equation}
    \mathbf{U}^{(l)} = \text{GNN}( \mathbf{A}^{(*)}, \mathbf{U}^{(l-1)};\mathbf{W}^{(l)})\text{,}
\end{equation}
where $\mathbf{U}^{(0)} = \mathbf{X}$
and $\mathbf{U}^{(L)}$ at the final layer $L$ is the output. $\mathbf{U}^{(L)}$ contains the necessary information of the downstream tasks. $\mathbf{W}^{(l)}$ is the parameter matrix of layer $l=1,2,...,L$.

If $\mathbf{A}^{(*)}$ is obtained by anchor-based probability metric learning technique, we need to employ tow-step message passing \cite{chen2020iterative}. The first step is formulated by \cref{eq:the first step} involving aggregating messages from nodes to anchor points to obtain anchor features $\mathbf{V}$. The second step is formulated by \cref{eq: the second step} involving aggregating messages from anchor points to nodes to update node embeddings $\mathbf{U}$.
\begin{equation}
    \mathbf{V}^{(l)}=\text{MP}_1(\mathbf{U}^{(l-1)},\mathbf{A}^{(*)}) = \Lambda^{-1} \mathbf{A}^{(*)^{\top}}\mathbf{U}^{(l-1)} \text{,}
    \label{eq:the first step}
\end{equation}
\begin{equation}
    \mathbf{U}^{(l)} = \text{MP}_2(\mathbf{V}^{(l)},\mathbf{A}^{(*)}) = \Delta^{-1}\mathbf{A}^{(*)} \mathbf{V}^{(l)} \text{,}
    \label{eq: the second step}
\end{equation}
where $\Lambda$ and $\Delta$ are diagonal matrices defined as $\Lambda_{kk} = \sum_{j'=1}^{n} \mathbf{A}^{(*)}_{j'k}$ and  $\Delta_{ii} = \sum_{k'=1}^{s} \mathbf{A}^{(*)}_{ik'}$ respectively. 
\subsection{Graph and Node Representation Joint Learning}
\label{chapter:loss}
The parameters within the graph learning module, i.e.,  $\mathbf{\Phi}$ and $t$, cannot be optimized solely through cross-entropy loss $\mathcal{L}_{pred}$.To optimize the graph learning, we introduce a graph loss, as described in \cite{kazi2022differentiable}, which rewards edges contributing to correct classification and penalizes edges leading to misclassification.

Let $\hat{\mathbf{y}} = (\hat{y}_1,\cdots,\hat{y}_i)$ denotes node labels predicted by our model and the vector of truth labels is denoted as $\mathbf{y}$. The graph loss function is as follows,
\begin{equation}
\mathcal{L}_{\mathcal{G}} = \sum_{i=1}^{N}\sum_{j \in \mathcal{N}_i(\mathbf{A})} \delta(y_i, \hat{y}_i) \log {s}_{ij}(\mathbf{\Phi}, t)
\end{equation}
where ${s}_{ij}$ is the element of $\hat{\mathbf{P}}$ or $\hat{\mathbf{R}}$. Additionally, $\delta\left(y_i, \hat{y}_i\right)$ denotes the reward function, indicating the difference between the average predicted accuracy and the current accuracy for node $i$. Specifically, $\delta(y_{i},\hat{y}_{i})=\mathbb{E}((c_{i}))-c_{i}$, where $c_i=1$ if $\hat{y}_i=y_i$ and 0 otherwise. 

Then we propose to jointly learning the graph structure and node representation by minimizing a hybrid loss function combining the cross-entropy loss $\mathcal{L}_{pred}$ and the graph loss $\mathcal{L}_{\mathcal{G}}$, namely, $\mathcal{L} = \mathcal{L}_{pred} + \mathcal{L}_{\mathcal{G}}$.

The \cref{alg:model} represents the computational procedure of the proposed model.

\section{Experiment}
In this section, we study the benefits of the PPGNN and compare it with state-of-the-art methods on node classification task. 

\subsection{Datasets and Setup}
We use four popular graph datasets: \texttt{Cora}, \texttt{PubMed}, \texttt{CiteSeer} \cite{sen2008collective}, and \texttt{Photo} \cite{shchur2018pitfalls}. The first three, \texttt{Cora}, \texttt{PubMed}, and \texttt{CiteSeer}, are citation datasets, where nodes represent scientific publications described by word vectors, and edges denote citation relationships between nodes. On the other hand, \texttt{Photo} is segments of the Amazon co-purchase graph \cite{mcauley2015image}, where nodes represent goods described by bag-of-words encoded product reviews, and edges indicate items that are frequently bought together. Please refer to \cref{details of datasets} for detailed statistics of datasets.

In this experiment, we focus on the transductive node classification task, where all nodes are observed during training but only the train set has labels.
For our model,
GCN is used as aggregate function with three layers having hidden dimensions of 64. Training involves the Adam optimizer with a learning rate set at $5 \times 10^{-3}$. The train/validation/test splits and all other settings follow \cite{kazi2022differentiable}. For each dataset, we execute 5 runs with different random seeds and report the mean accuracy and standard deviation. All our experiments are conducted on the NVIDIA RTX 3090 24GB GPU.

\begin{table}[ht]
    \centering
    \caption{Summary of datasets }
    \setlength{\tabcolsep}{25pt}
    \scalebox{0.9}{
        \begin{tabular}{ccccccc}
            \toprule
            &  Cora & CiteSeer & PubMed & Photo \\
            \midrule
             \# Nodes & 2708 & 3327 & 19717 & 7650\\
             \# Edges & 5278 & 4552 & 44324 & 119081\\
             \# Features & 1433 & 3703 & 500 & 745\\
            \# Classes & 7 & 6 & 3 & 8\\
            Average Degree & 3.9 & 2.7 & 4.5 & 31.1\\
            \bottomrule
        \end{tabular}
    }
    \label{details of datasets}
\end{table}

\subsection{Baselines}
To evaluate our method, we consider some baselines 
as follows,
\begin{enumerate}
    \item MLP (Multi-layer Perceptron) : MLP neglects the graph structure.
    \item Classical GNNs.
    \begin{itemize}
        \item GCN (Graph Convolutional Network) \citep{kipf2017semi}: GCN performs a simple diffusion operation over node features;
        \item GAT (Graph Attention Network) \citep{velivckovic2018graph}: GAT refines the diffusion process by learning per-edge weights through an attention mechanism.
    \end{itemize}
    \item Latent graph inference models that only accept node features as the input.
    \begin{itemize}
        \item $k$NN-GCN: $k$NN-GCN constructs a $k$NN (sparse $k$-nearest neighbor) graph based on node feature similarities and subsequently feeds it into GCN;
        \item SLAPS \cite{fatemi2021slaps}: SLAPS provides more supervision for inferring a graph structure through self-supervision.
    \end{itemize}
    \item Latent graph inference models that accept node features  and original graph structure.
    \begin{itemize}

        \item LDS \cite{franceschi2019learning}: LDS jointly learns the graph structure and parameters of a GCN;
        \item IDGL \cite{chen2019deep} : IDGL jointly and iteratively learns graph structure and graph embedding
        \item IDGL-ANCH \cite{chen2020iterative} : IDGL-ANCH is a variant of IDGL, which reduces time complexity through anchor-based approximation \cite{liu2010large};
        \item dDGM \cite{kazi2022differentiable}: dDGM is a learnable function that predicts a sparse adjacency matrix which is optimal for downstream task.
        We use Euclidean and hyperbolic space geometries for the graph embedding space with GCN as the aggregation function, denoted as dDGM-E and dDGM-H, respectively;
        \item dDGM-EHH and dDGM-SS \cite{de2022Latent}:  dDGM-EHH and dDGM-SS incorporates Riemannian geometry into the dDGM, representing embedding spaces with a torus and a manifold of Euclidean plane and two hyperboloids.
    \end{itemize}
\end{enumerate}

\subsection{Performance}

\begin{table}[!ht]
\centering
\caption{Results of accuracy on nodes classification task for the baselines and the proposed method. We report the mean and standard deviation (in percent) of accuracy on 5 runs for MLP, GCN, GAT and our model, as well as all for \texttt{Photo}. The others are obtained from the respective official reports. OOM indicates out of memory, OOT indicates out of time.}
\setlength{\tabcolsep}{11pt}
\begin{tabular}{ccccc}

\toprule
Methods & Cora & CiteSeer & PubMed & Photo\\
\midrule
MLP & $62.98_{ \pm 2.624}$ & $65.06_{ \pm 3.469}$ & $85.26_{ \pm 0.633}$ &$69.60_{ \pm 3.800}$ \\
GCN & $78.74_{ \pm 1.250}$ & $67.74_{ \pm 1.723}$ & $83.60_{ \pm 1.233}$ &$92.82_{ \pm 0.653}$ \\
GAT & $80.10_{ \pm 1.672}$ & $67.74_{ \pm 1.723}$ & $82.56_{ \pm 1.436}$ &$91.80_{ \pm 1.428}$ \\
\midrule
$k$NN-GCN & $66.50_{ \pm 0.400}$ & $68.30_{ \pm 1.300}$ & $70.40_{ \pm 0.400}$ & $78.28_{ \pm 1.676}$\\

SLAPS & $74.20_{ \pm 0.500}$ & $73.10_{ \pm 1.000}$ & $74.30_{ \pm 1.400}$&$46.72_{ \pm 0.110}$ \\
\midrule
LDS & $84.08_{ \pm 0.400}$ & $75.04_{ \pm 0.400}$ &  OOT  &OOT\\
IDGL & $84.50_{ \pm 0.300}$ & $74.10_{ \pm 0.200}$ &  OOM &$90.13_{ \pm 0.200}$ \\
IDGL-ANCH & $84.40_{ \pm 0.200}$ & $72.00_{ \pm 1.000}$ & $83.00_{ \pm 0.200}$&$87.60_{ \pm 0.320}$ \\

dDGM-E & $84.60_{ \pm 0.852}$ & $74.80_{ \pm 0.924}$ & $87.60_{ \pm 0.751}$&$93.06_{ \pm 0.670}$ \\
dDGM-H& $84.40_{ \pm 1.700}$ & $74.60_{ \pm 0.763}$ & $86.60_{ \pm 0.952}$& $91.48_{ \pm 2.871}$\\
dDGM-EHH & $86.63_{ \pm 3.250}$ & $75.42_{ \pm 2.390}$ & $39.93_{ \pm 1.350}$&$--$ \\
dDGM-SS & $65.96_{ \pm 9.460}$ & $59.16_{ \pm 5.960}$ & $87.82_{ \pm 0.590}$&$--$ \\
\midrule
PPGNN  & $87.10_{ \pm 0.326}$ & $76.34_{ \pm 0.580}$ & $\bm{88.40_{ \pm 0.770}}$ &$\bm{93.40_{ \pm 0.604}}$\\
PPGNN-anchor &$\bm{88.60_{ \pm 0.960}}$ & $\bm{78.08_{ \pm 0.810}}$ & $74.08_{ \pm 2.07}$ &$90.60_{ \pm 0.450}$\\
\bottomrule
\end{tabular}

\label{tab:results of models}
\end{table}
\textbf{Results.}
The results of the baselines and the proposed model are reported in \autoref{tab:results of models}. The results of baselines, excluding MLP, GAT, GCN and the results on the \texttt{Photo}, are sourced from the respective official reports. We can see that our model consistently outperforms all baselines on four datasets, indicating the efficacy of the proposed method. 

Firstly, we compare MLP with $k$NN-GCN and SLAPS, which rely solely on node features. It is observed that $k$NN-GCN and SLAPS perform better than MLP on \texttt{Cora} and \texttt{CiteSeer} but worse on \texttt{PubMed} and \texttt{Photo}. This emphasizes the importance of graph structure, with a lower-quality graph structure negatively impacting model performance.

Further comparison of latent graph inference models involves those utilizing only node features ($k$NN-GCN, SLAPS) and those using both node features and the original graph (LDS, IDGL, dDGM). 
It is worth noting that LDS, IDGL, and dDGM incorporate the original graph as part of their input and achieve better performance than the methods that only utilize node features. This suggests that the collaborative use of both original graph structures and node features contributes to the performance.

We explore different graph embedding spaces within dDGM models (dDGM-E, dDGM-H, dDGM-EHH, dDGM-SS). Among them, dDGM-EHH performs best on \texttt{Cora} and \texttt{CiteSeer}, while dDGM-SS performs best on \texttt{PubMed}. This indicates that the choice of embedding space impacts the similarity measurement matrix and latent graph structures. To further demonstrate the superiority of our proposed PPGNN, we compare it with dDGM-E. We observe a significant performance improvement in our model. This indicates PPGNN can help to avoid learning the wrong latent graph structures when the depth is not enough. 

\begin{figure}[!ht]
    \centering

    \begin{minipage}[t]{1\textwidth}
        \centering     \includegraphics[width=0.8\textwidth]{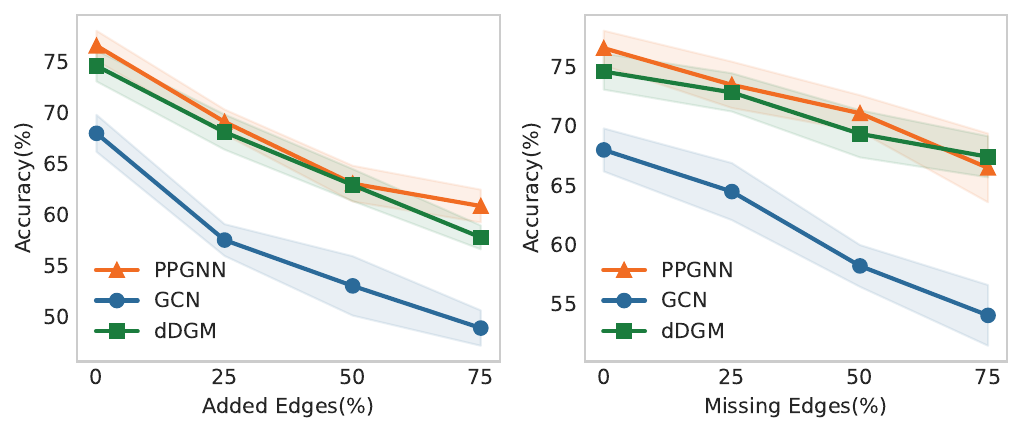}
        \subcaption{Cora}
    \end{minipage}

    \begin{minipage}[t]{1\textwidth}
        \centering
\includegraphics[width=0.8\textwidth]{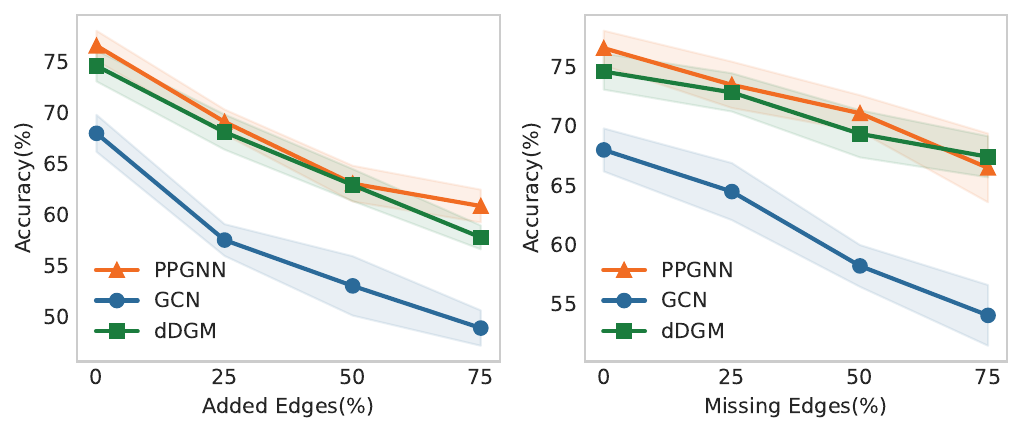}
\subcaption{CiteSeer}
    \end{minipage}
    
   \caption{Test accuracy ($\pm$ standard deviation ) in percent for the edge addition and deletion on Cora and CiteSeer.}
   \label{fig:noise}
\end{figure}
\begin{wrapfigure}{r}{0.45\textwidth}
  \centering
   \includegraphics[width=0.95\linewidth]{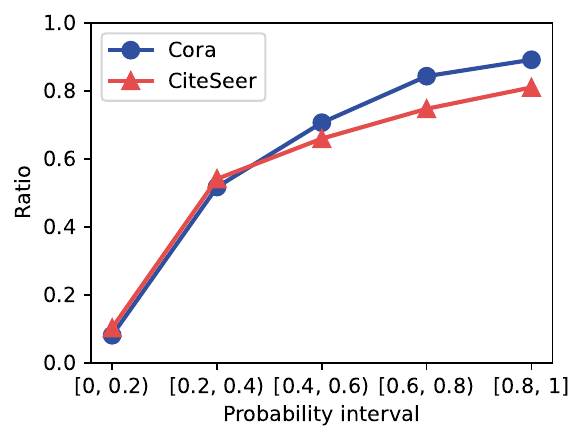}
   \caption{The ratio of two nodes in the test set sharing the same label in different probability interval.}
   \label{fig:odds}
\end{wrapfigure}
\textbf{Robustness.}
To assess the robustness of PPGNN, we construct graphs with random edges deletion and addition on \texttt{Cora} and \texttt{CiteSeer} as shown in \autoref{fig:noise} (a) and (b) respectively. Specifically,  we add or remove edges at ratios 25\%, 50\%, and 75\% to the existing edges for each dataset. Compared to GCN and dDGM, PPGNN achieves better results in both scenarios. Especially in scenarios involving edge addition, PPGNN consistently demonstrates outstanding performance across all ratios, showcasing its robust ability to eliminate noise from the graph.

\textbf{Homophily.}
We compute the ratio of the number of nodes pairs with the same label to the total number of nodes pairs in different probability intervals, using the test set of \texttt{Cora} and \texttt{CiteSeer}, as shown in \autoref{fig:odds}. We can see that the higher the probability of a connection between two nodes by \cref{eq:node probability}, the more likely it is that their labels are consistent. This indicates that PPGNN effectively captures and reinforces homophilic relationships in the graph structure.


\subsection{Model Analysis}

\subsubsection{Aggregate function}
In our analysis of aggregate functions, we consider three options: GCN \cite{kipf2017semi}, GAT \cite{veli2018graph}, and EdgeConv \cite{wang2019dynamic}. The results, as depicted in \cref{diffusion functions}, show comparable performance between GCN and GAT, with no significant difference. However, the adoption of EdgeConv notably reduces accuracy, particularly on small datasets.

This observation can be explained by the fact that a latent graph structure obtained through Probability Passing module already incorporates the information about nodes interactions. Therefore, GAT does not significantly contribute to performance improvement. Additionally, during the training process of our model, each sampled latent graph varies, making it challenging for the loss involving edge attributes to converge. Consequently, the use of EdgeConv leads a decrease in accuracy.

\begin{table}[ht]
\centering
\caption{Results of our model with different aggregate fuctions. We report the mean and standard deviation (in percent) of the accuracy on 5 runs.}
\setlength{\tabcolsep}{29pt}
\scalebox{0.9}{
\begin{tabular}{cccc}
\toprule
$\text{Aggre}(\cdot)$ & Cora & CiteSeer & PubMed \\
\midrule
GCN & $87.10_{ \pm 0.326}$ & $\mathbf{76.34_{ \pm 0.580}}$ & $\mathbf{88.40_{ \pm 0.770}}$ \\
GAT & $\mathbf{87.30_{ \pm 1.020}}$ & $76.22_{ \pm 1.130}$ & $88.20_{ \pm 0.704}$ \\
EdgeConv & $55.70_{ \pm 5.100}$ & $52.44_{ \pm 2.890}$ & $86.08_{ \pm 0.900}$ \\
\bottomrule
\end{tabular}
}
\label{diffusion functions}

\end{table}

\subsubsection{Time Complexity}
As for PPGNN, the cost of learning a probability matrix is $\mathcal{O}(N^2h)$ for $N$ nodes and embeddings in $\mathbb{R}^h$. Direct matrix multiplication in \cref{eq:node probability} has a computational complexity of $\mathcal{O}(N^3)$.
However, considering the sparsity of matrix $\mathbf{A}$, employing sparse matrix multiplication reduces the complexity to $\mathcal{O}(\| A^{(0)}\| N)$, where $\| A^{(0)}\|$ denotes the number of edges in the generated graph structure.  Hence the total complexity is $\mathcal{O}(N^2h+\| A^{(0)}\| N)$ in graph structure learning process.


As for PPGNN-anchor, the cost of learning a node-anchor probability is $\mathcal{O}(Nsh)$.While computing the result of Probability Passing in \cref{eq:anchor probability} costs $\mathcal{O}(N^2s)$. We can employ the sparse matrix multiplication to reduce the complexity to $\mathcal{O}(\| A^{(0)}\| s)$. The overall time complexity is $\mathcal{O}(Nsh+\| A^{(0)}\| s)$($s \ll n$).


\section{Conclusion}
\label{conclusin}
The graph structure is very important for GNNs. Many studies on latent graph inference have confirmed that is prevalent in popular graph datasets. The noise in generated graph tends to  be amplified during the message-passing process, impacting the performance of GNNs. To address this issue,  this paper introduces Probability Passing to improve performance. Our experimental results on four widely-used graph datasets for nodes classification task demonstrate the superior performance of our proposed model.

However, this method still has limitations. For instance, its computational complexity depends on the edges of the observed graph (if the observed graph has too many edges, it will affect computational efficiency). Additionally, this method can only be applied when the observed graph is available. It is not suitable for cases where the observed graph is unknown and where the generated graph structure from node features contains obvious noise.

In this study, we adopt the Probability Passing to correct the generated graph structure, which is similar to residual connection but is applied to non-Euclidean data. Besides that, it would be interesting to explore other methods to construct graph structure. We hope our work can inspire more researches into latent graph inference, able to infer graphs that are closer to the true underlying graph.

{
\bibliographystyle{plain}
\bibliography{ref}
}

\end{document}